\documentclass[12pt,table]{article}
\usepackage{graphicx} 
\usepackage{epstopdf}
\usepackage{amsthm,amsmath,amssymb, amsfonts, amscd, xspace, pifont}
\usepackage{epsfig, psfrag,pstricks}
\usepackage{times}
\usepackage[authoryear,round]{natbib}
\usepackage{algorithm, algorithmic}
\usepackage{booktabs}
\usepackage{multirow}
\usepackage{makecell}
\usepackage{array}
\usepackage{tabularx}

\newtheorem{proposition}{Proposition}
\providecommand{\keywords}[1]{\textbf{\textit{Keywords: }} #1}

\def\btheta{\mbox{\boldmath $\theta$}}

\def\bvarepsilon{\mbox{\boldmath $\varepsilon$}}

\def\bSigma{\mbox{\boldmath $\Sigma$}}

\def\bw{\mbox{\boldmath $\beta$}}

\def\bR{\mathbb{R}}
\def\mE{\mathbb{E}}
\def\bP{\mathbb{P}}
\def\mQ{\mathbb{Q}}

\def\bmu{\mbox{\boldmath $\mu$}}
\def\ba{{\bf a}}
\def\bb{{\bf b}}

\def\bm{{\bf m}}

\def\bQ{{\bf Q}}

\def\bw{{\bf w}}

\def\bu{{\bf u}}

\def\bI{{\bf I}}

\def\bx{{\bf x}}
\def\by{{\bf y}}
\def\bz{{\bf z}}

\def\nn{\nonumber}

\def\v2{\vspace{0.2in}}

\numberwithin{equation}{section}
\begin{document}
\setcounter{page}{1}
\baselineskip=15pt
\footskip=.3in
\parskip=5pt

\title{Schr\"odinger bridge based deep conditional generative learning}

\date{}

\author{Hanwen Huang \\\\
    {\it Department of Biostatistics, Data Science and Epidemiology}\\
    {\it Medical College of Georgia}\\
    {\it Augusta University, Augusta, GA 30912}}

\maketitle

\begin{abstract}
Conditional generative models represent a significant advancement in the field of machine learning, allowing for the controlled synthesis of data by incorporating additional information into the generation process. In this work we introduce a novel Schr\"odinger bridge based deep generative method for learning conditional distributions. We start from a unit-time diffusion process governed by a stochastic differential equation (SDE) that transforms a fixed point at time $0$ into a desired target conditional distribution at time $1$. For effective implementation, we discretize the SDE with Euler-Maruyama’s method where we estimate the drift term nonparametrically using a deep neural network. We apply our method to both low-dimensional and high-dimensional conditional generation problems. The numerical studies demonstrate that though our method does not directly provide the conditional density estimation, the samples generated by this method exhibit higher quality compared to those obtained by several existing methods. Moreover, the generated samples can be effectively utilized to estimate the conditional density and related statistical quantities, such as conditional mean and conditional standard deviation.

\end{abstract}
 

\section{Introduction}

Generative models are a powerful class of machine learning algorithms designed to generate new data samples that resemble a given dataset. These models learn the underlying distribution of the data and can produce realistic samples in various domains such as image synthesis, text generation, and speech synthesis \citep{sohldickstein2015deep,YangSong2021,ho2020denoising,song2021maximum,huang2022prodiffprogressivefastdiffusion,kingma2023variational}. However, standard generative models operate in an unsupervised manner, generating outputs without specific guidance or conditions. This is where conditional generative models come into play. 

Conditional generative models extend the capabilities of traditional generative models by incorporating additional information or conditions into the generation process. These conditions can be labels, class information, or any auxiliary data that influences the output. By conditioning the generation process, these models can produce more targeted and relevant outputs, making them highly valuable for applications that require controlled synthesis of data. For examples, in image synthesis, conditional models can generate images with specific attributes, such as generating pictures of animals with certain characteristics. In text generation, they can produce contextually relevant responses or translate text from one language to another with high fidelity. In the medical field, conditional generative models can be used to generate realistic medical images based on specific patient data, aiding in diagnosis and research.

To generate conditional samples, an intuitive strategy is to first estimate the conditional distribution, then generate conditional samples using different sampling schemes such as direct sampling or Markov Chain Monte Carlo. Various parametric or non-parametric approaches have been proposed in the literature. Examples include the smoothing method \citep{hallyao,Izbicki}, regression reformulation \citep{fan2004}, copula-based estimation \citep{jaworski2010copula}, conditional kernel density estimation (CKDE \cite{fan2004}), nearest neighbor conditional density estimation (NNKCDE \cite{dalmasso2020conditional}), flexible conditional density estimator (FlexCode \cite{izbicki2017converting}), and many others. All these methods focus on either univariate density estimation or multivariate density estimation with low dimensions. It is difficult to generalize them to high-dimensional settings because as the number of dimensions increases, the traditional multivariate density estimations are usually computationally expensive and moreover the amount of data needed to get a good density estimate grows exponentially. 

Recently, diffusion based generative models have been applied to the problem of conditional sampling generation and gained wide appraisal for their performance. The advantage of the diffusion based methods is that they are able to estimate the samplers directly without needing to know the underlying density and thus overcome the notorious problem of high-dimensional density estimation. Score-based generative models (SGM) are a class of generative models that have become recently very popular. The goal of SGM is to find nonlinear functions that transfer simple distributions into data distributions through first diffusing data to noise and then learning the score function to reverse this diffusion process. SGMs have been extended to conditional simulation case and achieved state-of-the-art performance in many application, see e.g. \cite{karras2019,saharia2021,tashiro2021}. However, performing conditional simulation using SGMs is computationally expensive as one needs to run the forward noising diffusion long enough to converge to the reference distribution. To address this issue, a line of recent works \citep{debortoli2023diffusion,vargas2022bayesian,chen2023likelihood} inspired by Schrödinger bridge (SB) has been proposed to accelerate simulation time which is complementary to many other acceleration techniques such as knowledge distillation \citep{luhman2021},  non-Markovian forward process and subsampling \citep{song2022}, and optimized noising diffusion \citep{watson2022}. The SB solution is an optimal diffusion process that transforms between two arbitrary distributions in a finite time horizon. The use of the SB formulation has been developed in the context of conditional simulation in \cite{shi2022}. By adapting the diffusion SB algorithm of \cite{debortoli2023diffusion}, an iterative algorithm called conditional diffusion SB (CDSB), was proposed to approximate the solution to the conditional SB problem. The idea is to employ iterative solvers which alternate between forward and backward updates, gradually refining the transport plan. Each iteration adjusts the process to better match the initial and target distributions while incorporating the conditional information.

CDSB is deemed as computationally demanding because at each iteration a deep neural network needs to be trained which depends on the results obtained in the previous steps. Consequently, it requires the storage of an increasing number of neural networks, two per iteration. \cite{huang2024} proposed an efficient method to generate unconditional samples based on a tractable class of SB diffusion processes that transport a Dirac delta distribution to the target distribution. The solutions of this SB system admit analytic forms which solely rely on the given data samples. This approach is a one-step solution which does not need to run iteration. Moreover, it does not require training neural network and thus substantially reduces the computation burden. In this paper, we extend this method from unconditional case to conditional case.

{\bf Related Work:} \cite{jiaocjasa} introduce a deep generative approach called generative conditional diffusion sampler (GCDS) extending the standard generative adversarial networks (GAN) framework \citep{goodfellow2014} by conditioning both the generator and the discriminator on auxiliary information. While GCDS succeeds in producing high quality samples, they are hard to train due to training instabilities, mode collapse, and necessitating considerable human tuning  \citep{karras2019}. Conditional variational autoencoders (cVAE) \citep{Sohn2015} were proposed as an extension to the VAE \citep{kingma2013auto} and allowed for the generation of data adhering to the specified condition. \cite{winkler2023} present conditional normalizing flows (CNF) to model the distribution of high-dimensional output as a conditional generative model. In CNF, complicated distributions are modeled by transforming a simple base density through the change of variables formula.  cVAE and CNF are less computationally expensive but tend to produce samples of lower visual quality. Moreover, normalizing flows put restrictions on the model architectures, thus limiting their expressivity \citep{batzolis2021}. In the framework of SGM, classifier guidance was proposed for training a conditional score network \citep{dhariwal2021}, which applies with discrete guidance, such as class labels. \cite{ho2022} developed classifier-free guidance to allow both discrete and continuous guidance. \cite{batzolis2021} introduced a multi-speed diffusion framework which leads to conditional multi-speed diffusive estimator (CMDE), where different parts of the input tensor diffuse according to different speeds. Recently, \cite{chang2024} introduced an ordinary differential equation (ODE) based deep generative method for learning conditional distributions, called conditional F\"ollmer flow (CFF). The idea is to first build a special SDE-based diffusion process, which can transform the target conditional density into a standard Gaussian density. Then the algorithm was implemented by discretizing the flow of the associated ODE with Euler’s method after estimating the velocity field using a deep neural network. CFF shared similar spirit to our proposed method in the sense that it is simulation free and its diffusion process transforms simple distribution into target distribution at finite time. The difference is that the diffusion process in our framework starts from a fixed point instead of a standard Gaussian. Moreover, our framework has more flexibility in selecting the reference SDE which can come from a more general class that subsumes the one used in CFF as a special case. 

The rest of the paper is organized as follows: Section \ref{sec:sbp} reviews the Schr\"{o}dinger bridge problem along with its application in unconditional sample generation. In Section \ref{condsb}, we elucidate how to employ the Schr\"{o}dinger bridge technique to design numerical algorithms for conditional sampling. Section \ref{numerical} presents numerical experiments to demonstrate the performance of the proposed method in both simulated and real datasets. Finally, in Section \ref{conclusion}, we discuss our work and outline some future directions. All technical proofs are provided in the Appendix. 

\section{Data-driven Schr\"odinger bridge sampler}\label{method}

Our goal is to learn the underlying distribution of i.i.d. samples $\{\bx_i\}_{i=1}^n\in\bR^{d}$, enabling it to generate new data points that resemble the original dataset. The method is rooted in the stochastic diffusion framework which is formulated by a Schr\"{o}dinger bridge (SB) problem aiming at transforming one probability distribution into another over a finite time horizon. Below, in Section \ref{sec:sbp}, we first provide an exposition of the background related to SB problems. In Section \ref{sec:uncond}, we present how to use SB to generate samples from unconditional distributions.

\subsection{Introduction to Schr\"{o}dinger bridge}\label{sec:sbp}

The SB problem \citep{leonard2013,ChenOptimal,debortoli2023diffusion,liu2023schrodinger} seeks the most likely evolution of a multivariate probability distribution from an initial state $\mu_0$ to a terminal state $\mu_1$, subject to stochastic dynamics. Lets first recall how the SB problem was applied to perform unconditional simulation. Denote $\bP$ and $\mQ$ the two probability measures over the functional space $C([0,1])$ constituted of continuous functions on the time interval $[0,1]$. Denote $\mQ_t$ the marginal distribution of $\mQ$ at time $t$. Given $\bP$, the The SB problem bridge problem seeks to find a $\mQ$ which has the closest distance to $\bP$ subject to the boundary constraints $\mQ_0=\mu_0$ and $\mQ_1=\mu_1$. It can be formulated in terms of the following optimization problem
\begin{eqnarray}\label{kldis}
\min_{\mQ\in\mQ_{\mu_0,\mu_1}}KL(\mQ|\bP),
\end{eqnarray}
where $\mQ_{\mu_0,\mu_1}=\{\mQ:\mQ_0=\mu_0,\mQ_1=\mu_1\}$ and $KL(\mQ|\bP)$ is the Kullback–Leibler distance between $\mQ$ and $\bP$ which is defined as
\begin{eqnarray}\label{kldef}
D(\mQ\|\bP)&=&\left\{\begin{array}{cc}\int\log(d\mQ/d\bP)d\mQ&if~\mQ\ll\bP\\
\infty&otherwise\end{array}\right.,
\end{eqnarray}
where $\mQ\ll\bP$ denotes that $\mQ$ is absolutely continuous w.r.t. $\bP$ and $d\mQ/d\bP$ represents the Radon-Nikodym derivative of $\mQ$ w.r.t. $\bP$. One usually chooses simple reference measure $\bP$ which is governed by a stochastic differential equation (SDE) 
\begin{eqnarray}\label{sde0}
d\bx_t=\bb(\bx_t,t) dt+\sigma(t) d\bw_t,~\bx_0\sim\mu_0,
\end{eqnarray}
where $\bx_t$ is a $d$-dimensional stochastic process indexed by $t\in[0,1]$, $\bw_t$ is a $d$-dimensional standard Brownian motion, $\bb(\bx,t):\bR^d\times[0,1]\rightarrow\bR^d$ and $\sigma(t):[0,1]\rightarrow(0,\infty)$ are drift and diffusion respectively. In this paper, we consider two reference SDEs
\begin{eqnarray}\label{ref1}
d\bx_t&=&\sqrt{\alpha^\prime(t)} d\bw_t,\\\label{ref2}
d\bx_t&=&-\frac{1}{2}\beta(t)\bx_t dt+\sqrt{\beta(t)} d\bw_t,
\end{eqnarray}
where $\alpha(t)$ and $\beta(t)$ are non-negative functions of $t\in[0,1]$. Since both (\ref{ref1}) and (\ref{ref2}) have affine drift coefficients, their perturbation kernels $\pi(\bx_t|\bx_0)$ are all Gaussian and can be computed in closed-forms. In particular, if $\alpha(t)=1$, (\ref{ref1}) is the $d$-dimensional standard Brownian motion; if $\beta(t)=1$, (\ref{ref2}) is the Ornstein–Uhlenbeck process. As shown in \cite{song2021scorebased}, the discretizations of (\ref{ref1}) and (\ref{ref2}) correspond to the two successful classes of score based generative models: score matching with Langevin dynamics and denoising diffusion probabilistic modeling, respectively.

The SB problem (\ref{kldis}) can also be viewed as a entropy-regularized optimal transport problem between $\mu_0$ and $\mu_1$ with regularized transportation cost \citep{leonard2013,ChenOptimal,Peluchetti2023}. Were $\mQ$ available, we would obtain a generative model by ancestral sampling: starting from $\bx_0=\mu_0$, then sample $\bx_{k+1}\sim\pi(\bx_{k+1}|\bx_k)$ for $k\in\{0, \cdots, N\}$. However, the SB problem usually does not admit a closed-form solution for $\mQ$. It is often solved numerically using Iterative Proportional Fitting algorithms that alternate between forward and backward updates of the corresponding half bridge problems \citep{Kullback1968,debortoli2023diffusion,vargas2022bayesian}, akin to the Sinkhorn algorithm used in optimal transport \citep{sinkhorn}. 

\subsection{Schr\"{o}dinger bridge based generative learning}\label{sec:uncond}

In special case where the initial state follows a Dirac Delta distribution, i.e. $\mu_0=\delta_{\ba}$, where $\ba\in\bR^d$, the problem (\ref{kldis}) has a closed form solution \citep{Pavon1989,DaiPra1991,pavon2018datadriven}. The Theorem 1 in \cite{huang2024} states that the solution $\mQ$ to (\ref{kldis}) is governed by the following SDE  
\begin{eqnarray}\label{sde}
d\bx_t=[\bb(\bx_t,t)+\bu^\star(\bx_t,t)] dt+\sigma(t) d\bw_t,~\bx_0=\ba,
\end{eqnarray}
where, in contrast to (\ref{sde0}), the drift function includes an extra term $\bu^\star(\bx,t):\bR^d\times[0,1]\rightarrow\bR^d$ which is given by
\begin{eqnarray}\label{drift0}
\bu^\star(\bx,t)&=&\frac{\sigma(t)^2\int\nabla_{\bx}g_t(\bx,\bx_1)\mu_1(d\bx_1)}{\int g_t(\bx,\bx_1)\mu_1(d\bx_1)},
\end{eqnarray}
where $\mu_1(\bx_1)$ is the target density and
\begin{eqnarray}\label{transition}
g_t(\bx,\bx_1)&=&\frac{q(t,\bx,1,\bx_1)}{q(0,\ba,1,\bx_1)},
\end{eqnarray}
with $q(t_1,\bx,t_2,\by)$ denotes the transition density of $\bx_{t_2}=\by$ at time $t_2$ given $\bx_{t_1}=\bx$ at time $t_1$ for the stochastic process $\bx_t$ governed by the reference SDE (\ref{sde0}).

The result (\ref{drift0}) has been applied to the unconditional generative learning in \cite{huang2024}. The advantage of (\ref{drift0}) over other diffusion based methods  is that,  by appropriately choosing the reference SDEs, the drift term can be analytically determined and thus avoid the complexity of neural network training procedure. In order to extend it to conditional generative learning, we reformulate (\ref{drift0}) for the two special reference SDEs (\ref{ref1}) and (\ref{ref2}) in the following Proposition:
\begin{proposition}\label{thm1}
For reference SDE (\ref{ref1}), the drift term $\bu^\star(\bx_t,t)$ in (\ref{sde}) is a time varying vector field $\bu(\bx_t,t)$ that minimizes the following quadratic objective function
\begin{eqnarray}\label{lse01}
\mE_{\bQ}\left\|\frac{\alpha^\prime(t)}{\alpha(1)-\alpha(t)}(\bx_1-\bx_t)-\bu(\bx_t,t)\right\|^2,
\end{eqnarray}
where $\bQ=[t\sim{\cal U}(0,1)]\otimes\mu_1(\bx_1)\otimes\pi(\bx_t|\bx_1)$, and the conditional distribution $\pi(\bx_t|\bx_1)$ is defined through $\pi(\bx_t|\bx_1)\sim N(\bmu_{\bx_t|\bx_1},\sigma^2_{\bx_t|\bx_1})$, where
\begin{eqnarray}\nn
\bmu_{\bx_t|\bx_1}&=&\frac{\alpha(t)-\alpha(0)}{\alpha(1)-\alpha(0)}\bx_1+\frac{\alpha(1)-\alpha(t)}{\alpha(1)-\alpha(0)}\ba,\\\nn
\sigma^2_{\bx_t|\bx_1}&=&\frac{[\alpha(t)-\alpha(0)][\alpha(1)-\alpha(t)]}{\alpha(1)-\alpha(0)}.
\end{eqnarray}
For reference SDE (\ref{ref2}), the drift term $\bu^\star(\bx_t,t)$ minimizes the following objective
\begin{eqnarray}\label{lse02}
\mE_{\bQ}\left\|\frac{\beta(t)\xi}{1-\xi^2}(\bx_1-\xi\bx_t)-\bu(\bx_t,t)\right\|^2,
\end{eqnarray}
where the expectation is with respect to $\bQ=[t\sim{\cal U}(0,1)]\otimes\mu_1(\bx_1)\otimes\pi(\bx_t|\bx_1)$ and the conditional distribution of $\bx_t$ given $\bx_1$ is $\pi(\bx_t|\bx_1)\sim N(\bmu_{\bx_t|\bx_1},\sigma^2_{\bx_t|\bx_1})$, where
\begin{eqnarray}\nn
\bmu_{\bx_t|\bx_1}&=&\frac{\sigma_2^2}{\xi\sigma^2}\bx_1+\frac{\tau\sigma_2^2}{\xi\sigma_1^2}\ba,\\\nn
\sigma^2_{\bx_t|\bx_1}&=&\frac{(\sigma_2^2)^2}{\xi^2\sigma^2},
\end{eqnarray}
where 
\begin{eqnarray}\nn
\sigma_1^2&=&1-\tau^2,\\\nn
\sigma_2^2&=&1-\xi^2,\\\label{coef}
\sigma^2&=&\left(\frac{1}{\sigma_2^2}-\frac{1}{\sigma_1^2}\right)^{-1},\\\nn
\xi&=&e^{-\frac{1}{2}\int_t^1\beta(s^\prime)ds^\prime},\\\nn
\tau&=&e^{-\frac{1}{2}\int_0^1\beta(s^\prime)ds^\prime}.
\end{eqnarray}
\end{proposition}    
Equations (\ref{lse01}) and (\ref{lse02}) allow us to design alternative deep learning algorithms to estimate $\bu^\star(\bx_t,t)$ nonparametrically via working with a set of  i.i.d. samples $t_i\sim{\cal U}(0,1)$, $\bx_{1,i}\sim\mu_1(\bx_1)$, and $\bx_{t,i}\sim N(\bmu_{\bx_{t_i}|\bx_{1,i}},\sigma^2_{\bx_{t_i}|\bx_{1,i}})$. 

\section{Conditional Schr\"{o}dinger Bridge sampler}\label{condsb}

A conditional diffusion model is a modification of an unconditional diffusion model. By incorporating the SB approach into conditional generative models, we aim to improve the quality and robustness of generated samples conditioned on the auxiliary information which could be the discrete class label or continuous covariates. The corresponding SB process models the evolution of the stochastic process which transfers the initial distribution into the target distribution over a finite time horizon while respecting the conditional information. 

\subsection{Integrating Schr\"{o}dinger bridge with conditional generative models}

Suppose that we have a random vector $\bx\in\bR^{d_x}$ and a condition variable $\bz\in\bR^{d_z}$ related to $\bx$, forming a pair $(\bx,\bz)$. The marginal densities of $\bx$ and $\bz$ are denoted as $\mu_x(\bx)$ and $\mu_z(\bz)$, respectively, while the joint density of $(\bx,\bz)$ is denoted as $\mu_{x,z}(\bx,\bz)$. Using $\mu_{x^z}(\bx^{\bz})$ to represent the conditional density of $\bx$ given $\bz$, our interest lies in efficiently sampling from $\mu_{x^z}(\bx^{\bz})$.

Similar to the unconditional case, to sample from the conditional target distribution for any realization $\bz$, we start from a SB problem which solves the following minimization problem:
\begin{eqnarray}\label{kldiscond}
\mQ^{z\star}&=&\text{argmin}_{\mQ^{\bz}\in\mQ_{\ba,\mu_{x^z}(\bx^{\bz})}}KL(\mQ^{\bz}|\bP),
\end{eqnarray}
i.e. we aim to find a probability measure $\mQ^{z\star}$ which has the smallest KL distance to $\bP$ and also satisfies the boundary conditions: $\mQ^{\bz}_0=\ba$ and $\mQ^{\bz}_1=\mu_{x^z}(\bx^{\bz})$. However, the SB problem (\ref{kldiscond}) is not applicable in practice because it requires samples from $\mu_{x^z}(\bx^{\bz})$ for a given $\bz$ which is usually not available especially for continuous $\bz$. Here we propose instead to solve the following modified problem
\begin{eqnarray}\label{modifiedcond}
\tilde{\mQ}^{\star}&=&\text{argmin}_{\tilde{\mQ}\in{\cal Q}}\left[\mE_{\bz\sim\mu_z}\{KL(\mQ^{\bz}|\bP)\}:\mQ^{\bz}_0=\ba,\mQ^{\bz}_1\otimes\mu_z=\mu_{x,z}\right],
\end{eqnarray}
where ${\cal Q}=(\mQ^{\bz})_{\bz\in\bR^{d_z}}$ defines a measure over the functional space $C([0,1])$ for each $\bz$. This corresponds to an averaged version of (\ref{kldiscond}) over the distribution $\mu_z$. 

To convert an unconditional diffusion model into a conditional diffusion model, the corresponding SDE can be conditioned on a value $\bz$. The following Proposition presents the solution to the SB problem (\ref{kldiscond}) with respect to the two reference SDEs (\ref{ref1}) and (\ref{ref2}).
\begin{proposition}\label{thm2}
The solution $\mQ^{\bz}$ to the SBP (\ref{kldiscond}) is governed by the following SDE
\begin{eqnarray}\label{csde0}
d\bx_t^{\bz}=[\bb(\bx_t^{\bz},t)+\bu^\star(\bx_t^{\bz},\bz,t)]dt+\sigma(t) d\bw_t,~\bx_0^{\bz}=\ba,
\end{eqnarray}
where for the reference SDE (\ref{ref1}),  $\bu^\star:\bR^{d_x}\times\bR^{d_z}\times[0,1]\rightarrow\bR^{d_x}$ is a vector field that minimizes the following objective function
\begin{eqnarray}\label{lse1}
\mE_{\tilde{\bQ}}\left\|\frac{\alpha^\prime(t)}{\alpha(1)-\alpha(t)}(\bx_1^\bz-\bx_t^\bz)-\bu(\bx_t^\bz,\bz,t)\right\|^2,
\end{eqnarray}
where $\tilde{\bQ}=[t\sim{\cal U}(0,1)]\otimes\mu_{x,z}(\bx_1^{\bz},\bz)\otimes\pi(\bx_t^{\bz}|\bx_1^{\bz})$. Here the conditional distribution $\bx_t^{\bz}$ given $\bx_1^{\bz}$ is defined through $\pi(\bx_t^{\bz}|\bx_1^{\bz})\sim N(\bmu_{\bx_t^{\bz}|\bx_1^{\bz}},\sigma^2_{\bx_t^{\bz}|\bx_1^{\bz}})$, where
\begin{eqnarray}\nn
\bmu_{\bx_t^{\bz}|\bx_1^{\bz}}&=&\frac{\alpha(t)-\alpha(0)}{\alpha(1)-\alpha(0)}\bx_1^{\bz}+\frac{\alpha(1)-\alpha(t)}{\alpha(1)-\alpha(0)}\ba,\\\nn
\sigma^2_{\bx_t^{\bz}|\bx_1^{\bz}}&=&\frac{[\alpha(t)-\alpha(0)][\alpha(1)-\alpha(t)]}{\alpha(1)-\alpha(0)}.
\end{eqnarray}
Similarly, for the reference SDE (\ref{ref2}), $\bu^\star$ is determined by minimizing the following objective function
\begin{eqnarray}\label{lse2}
\mE_{\tilde{\bQ}}\left\|\frac{\beta(t)\xi}{1-\xi^2}\left[\bx_1^{\bz}-\xi\bx_t^{\bz}\right]-\bu(\bx_t^{\bz},\bz,t)\right\|^2,
\end{eqnarray}
where the expectation is with respect to ${\tilde{\bQ}}=[t\sim{\cal U}(0,1)]\otimes\mu_{x,z}(\bx_1^{\bz},\bz)\otimes\pi(\bx_t^{\bz}|\bx_1^{\bz})$ and the conditional distribution of $\bx_t^{\bz}$ given $\bx_1^{\bz}$ is $\pi(\bx_t^{\bz}|\bx_1^{\bz})\sim N(\bmu_{\bx_t^{\bz}|\bx_1^{\bz}},\sigma^2_{\bx_t^{\bz}|\bx_1^{\bz}})$, where
\begin{eqnarray}\nn
\bmu_{\bx_t^{\bz}|\bx_1^{\bz}}&=&\frac{\sigma_2^2}{\xi\sigma^2}\bx_1^{\bz}+\frac{\tau\sigma_2^2}{\xi\sigma_1^2}\ba,\\\nn
\sigma^2_{\bx_t^{\bz}|\bx_1^{\bz}}&=&\frac{(\sigma_2^2)^2}{\xi^2\sigma^2},
\end{eqnarray}
where the coefficients $\sigma_1^2,~\sigma_2^2,~\sigma^2,~\xi,~\tau$ are defined in (\ref{coef}).
\end{proposition}    
Proposition \ref{thm2} indicates that the dependence of the SB problem (\ref{kldiscond}) on $\bz$ is characterized by the dependence of the corresponding SDE drift term $\bu^\star$ on $\bz$. So the theoretical results for existence and uniqueness of the solution to the SB problem can be directly applied here.  For categorical $\bz$,  (\ref{kldiscond}) can be solved separately for each category by applying the method developed in \cite{huang2024} to the subset of samples for each specific category. For continuous $\bz$, the expressions in (\ref{lse1}) and (\ref{lse2}) motivate objectives for fitting $\bu^\star$ with neural network $\bu_{\btheta}(\bx,\bz,t)$ based on i.i.d. joint samples $\bx_1^{\bz},\bz\sim\mu_{x,z}(\bx,\bz)$, i.i.d. samples $t\sim{\cal U}(0,1)$, and conditional samples $\bx_t^{\bz}|\bx_1^{\bz}\sim N(\bmu_{\bx_t^{\bz}|\bx_1^{\bz}},\sigma^2_{\bx_t^{\bz}|\bx_1^{\bz}})$. Therefore, the problem of estimating the conditional distribution of $\bx$ given $\bz$ is now converted into a regression problem of estimating the drift term $\bu^\star$ as a function of $\bx$, $\bz$, and $t$. 
 
\subsection{Implementation}

In this section, we detail the implementation of using SB approach for conditional simulation, i.e. to sample from a distribution $\pi_{x^z}(\bx^{\bz})$ assuming only that it is possible to sample $\pi_{x,z}(\bx,\bz)$.

\subsubsection{Training process}

For reference SDE (\ref{ref1}), we treat the quadratic form (\ref{lse1}) as our training objective. In practice, (\ref{lse1}) is implemented using collected i.i.d. data points $\{(\bx_i, \bz_i)\}_{i=1}^n$, which essentially replaces the expectation over $(\bx,\bz)$ by its empirical counterpart. We denote a loss function
\begin{eqnarray}\label{lhat}
\hat{\cal L}(\btheta)&=&\frac{1}{mn}\sum_{i=1}^n\sum^m_{j=1}\left\|\frac{\alpha^\prime(t_j)}{\alpha(1)-\alpha(t_j)}(\bx^{(i)}-\bx^{(i)}_{t_j})-\bu_{\btheta}(\bx^{(i)}_{t_j},\bz^{(i)},t_j)\right\|^2,
\end{eqnarray}
where $(\bx^{(i)},\bz^{(i)})\sim p_{data}$,  $t_j\sim{\cal U}[\epsilon,1-\epsilon]$, and $\bx_{t_j}^{(i)}\sim N(\bmu_{\bx^{(i)}_{t_j}|\bx^{(i)}},\sigma^2_{\bx^{(i)}_{t_j}|\bx^{(i)}})$ with
\begin{eqnarray}\label{smean}
\bmu_{\bx^{(i)}_{t_j}|\bx^{(i)}}&=&\frac{\alpha(t_j)-\alpha(0)}{\alpha(1)-\alpha(0)}\bx^{(i)}+\frac{\alpha(1)-\alpha(t_j)}{\alpha(1)-\alpha(0)}\ba,\\\label{svar}
\sigma^2_{\bx^{(i)}_{t_j}|\bx^{(i)}}&=&\frac{[\alpha(t_j)-\alpha(0)][\alpha(1)-\alpha(t_j)]}{\alpha(1)-\alpha(0)}.
\end{eqnarray}
To circumvent potential numerical instability issues at $t = 0$ and $t = 1$, here we have introduced a truncation of the unit-time interval by $\epsilon$ at both endpoints, which is commonly adopted in practice \citep{YangSong2021,AlexNichol2021}. 

We fit $\bu_{\btheta}$ using a feed forward neural network which is expressive enough to approximate the true drift term. The training process of Schr\"{o}dinger Bridge based conditional generative learning (SBCG) is summarized in Algorithm \ref{alg1}. 
\begin{algorithm}[H]
	\caption{Training process of SBCG}
    \label{alg1}
	\begin{algorithmic}[1]
\STATE Input: input data $p_{data}$, number of iterations $L$.
\FOR{$l= 1,\ldots, L$ }
\STATE Sample $(\bx^{(i)},\bz^{(i)})\sim p_{data}$.
\STATE Sample $t_j\sim{\cal U}[\epsilon,1-\epsilon]$.
\STATE Sample $\bx^{(i)}_{t_j}\sim N(\bmu_{\bx^{(i)}_{t_j}|\bx^{(i)}},\sigma^2_{\bx^{(i)}_{t_j}|\bx^{(i)}})$ with $\bmu_{\bx^{(i)}_{t_j}|\bx^{(i)}}$ and $\sigma^2_{\bx^{(i)}_{t_j}|\bx^{(i)}}$ given in (\ref{smean}) and (\ref{svar}).
\STATE Compute $\hat{\cal L}(\btheta)$ with (\ref{lhat}).
 \STATE Update $\btheta$ with $\nabla_{\btheta}\hat{\cal L}(\btheta)$.
\ENDFOR
\STATE Output: $\bu_{\btheta}(\bx,\bz,t)$.
\end{algorithmic}
\end{algorithm}
The training processes of SBCG for reference SDE (\ref{ref2}) are similar to the scenario for reference SDE (\ref{ref1}), so we will not provide additional details here. Note that we choose to parametrize $\bu^\star(\bx,\bz,t)$ in (\ref{csde0}) by a neural network $\bu_{\btheta}(\bx,\bz,t)$ with trainable parameters $\btheta$. But our framework does not need to rely on neural networks; there can be other ways to represent the function $\bu_{\btheta}$ (e.g. kernel representation). For high-dimensional data, with sufficient training data, neural networks tend to have competitive performance due to their expressiveness power.

\subsubsection{Generative process}

After training, we can simply substitute the drift term $\bu^\star$ in (\ref{csde0}) with the learned network $\bu_{\btheta}$ to generate data from $\mu_{x^z}$ for a given $\bz$. In practice, a proper discretization is applied to generate samples, whose deviation to the continuous-time stochastic process can be controlled by the step size of the discretization \citep{chen2023}. We employ Euler-Maruyama’s method to discretize the SDE system in (\ref{csde0}) with a fixed step size. Let 
\begin{eqnarray}\nn
t_k=\epsilon+kh,~k=0,\cdots,K,~\text{with}~h=(1-2\epsilon)/K,
\end{eqnarray}
and set $\bx^\bz_0=\ba$. Then the Euler-Maruyama discretization of (\ref{csde0}) has the following form
\begin{eqnarray}\label{euler}
\bx_{k+1}^\bz=\bx_{k}^\bz+h[\bb(\bx_{k}^\bz,t_k)+\bu_{\btheta}(\bx_{k}^\bz,\bz,t_k)]+\sqrt{h}\sigma(t_k)\bvarepsilon_k,~k=0,\cdots,K-1,
\end{eqnarray}
where $\{\bvarepsilon_k\}^{K-1}_{k=0}$ are independent and identically distributed random vectors from $N(0,\bI_{d_x})$ and $\bu_{\btheta}$ is the output from Algorithm \ref{alg1}. Then the resulting sample $\bx_K$ will be approximately distributed from $\mu_{x^z}$ for a given $\bz$. The pseudocode for implementing (\ref{euler}) is presented in Algorithm \ref{alg2}. 
\begin{algorithm}[H]
	\caption{Generative process of SBCG}
    \label{alg2}
	\begin{algorithmic}[1]
\STATE Input: condition $\bz$, grid points $\epsilon=t_0\textless t_1\textless\cdots\textless t_K=1-\epsilon$ on time interval $[\epsilon,1-\epsilon]$ with step size $h=(1-2\epsilon)/K$, starting point $\bx_{t_0}^\bz=\ba$.
\FOR{$k= 0,1,\ldots, K-1$ }
\STATE Sample random vector $\bvarepsilon_{k}\sim N(0,\bI_{d_x})$,
\STATE Compute $\bu_{\btheta}(\bx_{t_k}^{\bz},\bz,t_k)$.
\STATE Update $\bx_{t_{k+1}}^\bz$ according to \eqref{euler}.
\ENDFOR
\STATE Output:  $\{\bx_{t_k}^\bz\}_{k=1}^{K}$
\end{algorithmic}
\end{algorithm}

Note that estimating the conditional density $\mu_{x^z}(\bx^\bz)$ is highly challenging in practice when $d_{x}$ and $d_{z}$ are large. Our framework has two main benefits. First, we can just implement Algorithms \ref{alg1} and \ref{alg2} to draw samples from $\mu_{x^z}(\bx^\bz)$ without estimating its density function. Second, compared to other popular conditional generative models such as GAN \citep{goodfellow2014,jiaocjasa} and conditional diffusion Schrödinger Bridges \citep{pmlr-v180-shi22a,shi2023diffusion}, our method is simpler and easier to train. This is because our objective (\ref{lse1}) only involves a single unknown function $\bu_{\btheta}(\bx,\bz,t)$ without additional discriminators to guide the training of it, and therefore no additional inner loops or iterations are required.

\section{Numerical experiments}\label{numerical}

In this section, we conduct numerical experiments to examine the performance of SBCG on both simulated and real data. First, in section \ref{sec:simu}, we carry out simulation studies to assess the effectiveness of our proposed method through the visulation of two-dimensional samples in section \ref{sec:visu} and the comparison of the estimated statistical quantities with the true values in section \ref{sec:mean}. Then, in section \ref{subsec:wine}, we use the wine quality dataset and the abalone data from UCI machine learning repository to demonstrate the advantages of the SBCG method in computing the prediction intervals for given predictors. In section \ref{subsec:mnist}, we study the high-dimensional conditional generation problems by applying our proposed method to the MNIST handwritten digits dataset.

\subsection{Simulation studies}\label{sec:simu}

\subsubsection{Visualization of two-dimensional examples}\label{sec:visu}

\begin{figure}[ht!]
    \vspace{0cm}
		\begin{minipage}[t]{0.31\linewidth}
\centering
\includegraphics[width=1.08\textwidth]{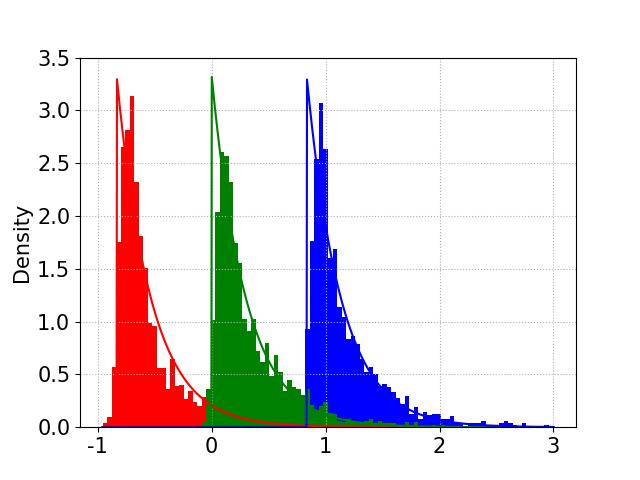}
		\end{minipage}
		\hspace{0.1cm}
		\begin{minipage}[t]{0.31\linewidth}
\centering
\includegraphics[width=1.08\textwidth]{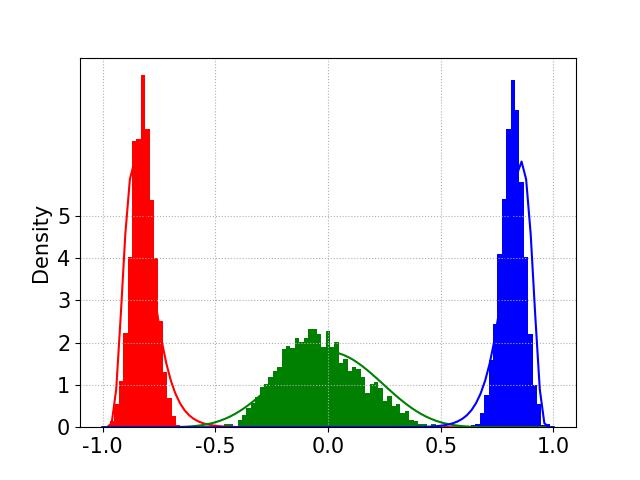}
		\end{minipage}
		\hspace{0.1cm}
		\begin{minipage}[t]{0.31\linewidth}
\centering
\includegraphics[width=1.08\textwidth]{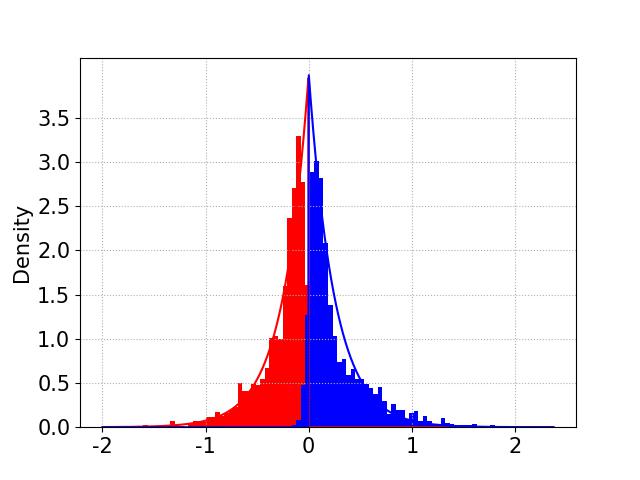}
		\end{minipage}
    \vspace{-0cm}
\caption{Sampling performance of SBCG on three two-dimensional datasets generated from examples (\ref{example1}). The conditional distributions of $x$  are sampled given $z=-1.2$ (red), $z=0$ (green), and $z=1.2$ (blue). The solid curves represent the corresponding true densities. From left to right: Example 1, Example 2, and Example 3.}
\label{figure1}
\end{figure}

A first test to benchmark the validity of the method is sampling a target density whose analytic form is known or whose density can be visualized for comparison. We first consider the two-dimensional examples of \cite{baptista2023conditionalsamplingmonotonegans} which include three nonlinear, non-Gaussian examples for $\mu_{x,z}(x,z)$. Given $z\sim\mu_z(z)={\cal U}[-3,3]$, the conditional distribution $\mu_{x^z}(x^z)$ is defined through
\begin{eqnarray}\nn
\text{Example 1}:&& x=\tanh(z)+\epsilon, ~\epsilon\sim\Gamma(1,0.3),\\\label{example1}
\text{Example 2}:&& x=\tanh(z+\epsilon), ~\epsilon\sim N(0,0.05),\\\nn
\text{Example 3}:&& x=\epsilon\tanh(z), ~\epsilon\sim\Gamma(1,0.3).
\end{eqnarray}
We run SBCG on each of the examples with 50,000 training points and use a neural network model with $K=100$ diffusion steps. The neural network model has four-layer fully connected ReLU network with hidden nodes 32, 64, 64, 32, respectively. We first train the drift estimator by matching on the training set through Algorithm \ref{alg1}, and then for a new $z$ we generate 2000 $x$ samples by Algorithm \ref{alg2} based on the trained drift estimator. Figure \ref{figure1} shows the resulting histogram of the learned $\mu_{x^z}$ and the true conditional probability density of $x$ given $z\in\{-1.2,0,1.2\}$. As can be observed, the SBCG method generates samples close to target.

\begin{figure}[ht!]
    \vspace{-0.5cm}
		\begin{minipage}[t]{0.23\linewidth}
\centering
\includegraphics[width=1.08\textwidth]{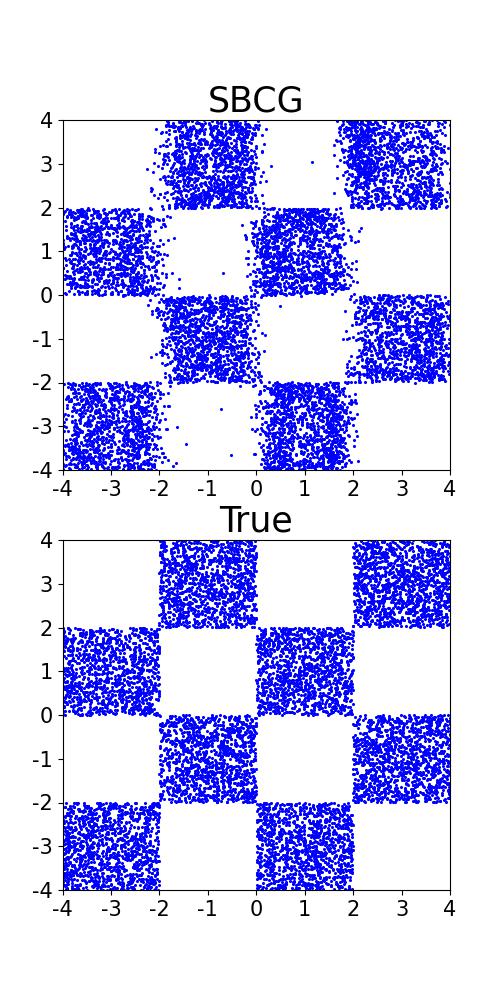}
		\end{minipage}
		\hspace{0.1cm}
		\begin{minipage}[t]{0.23\linewidth}
\centering
\includegraphics[width=1.08\textwidth]{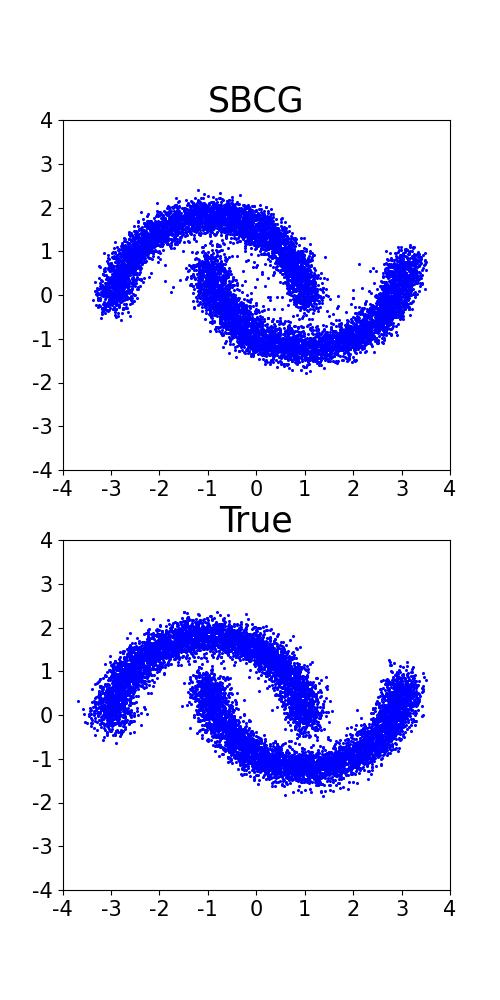}
		\end{minipage}
		\hspace{0.1cm}
		\begin{minipage}[t]{0.23\linewidth}
\centering
\includegraphics[width=1.08\textwidth]{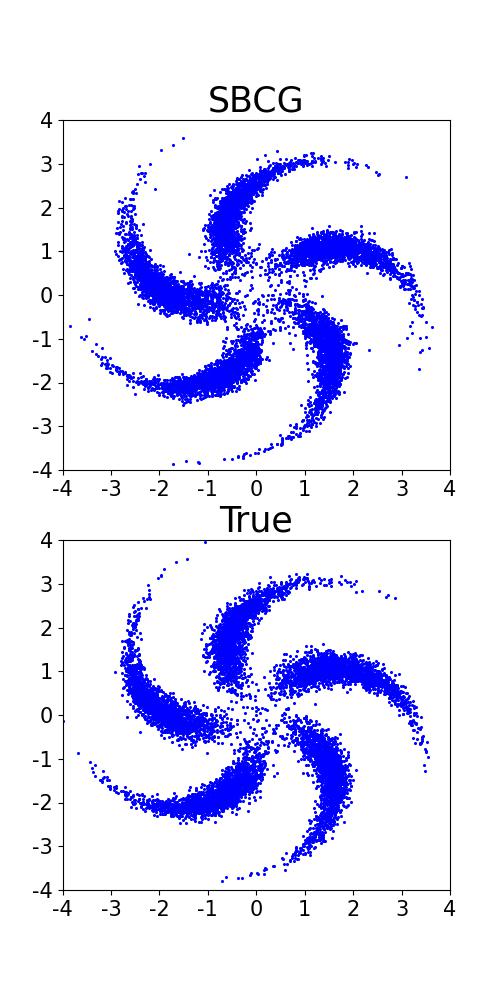}
		\end{minipage}
		\hspace{0.1cm}
		\begin{minipage}[t]{0.23\linewidth}
\centering
\includegraphics[width=1.08\textwidth]{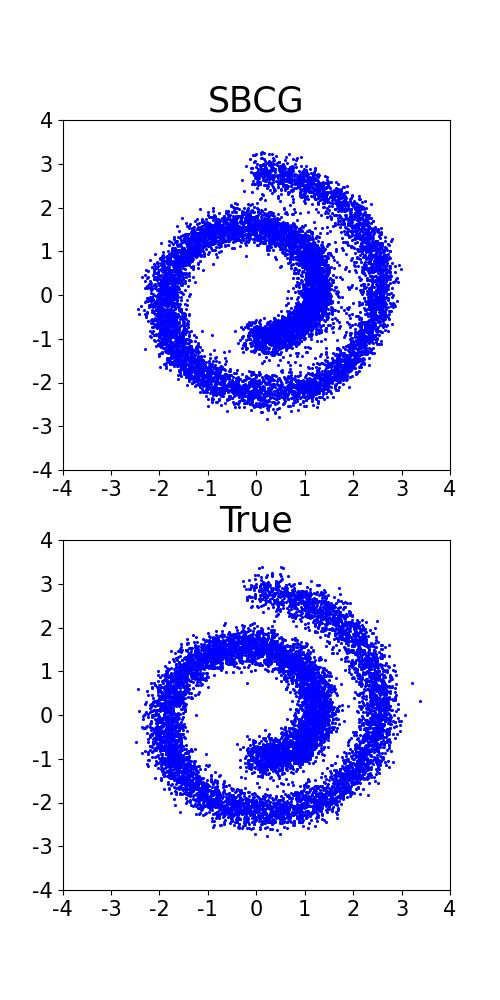}
		\end{minipage}
    \vspace{-0.5cm}
\caption{Scatter plots of joint distributions generated by SBCG and the ground truth. From left to right: checkerboard, moons, pinwheel, and swissroll. Bottom: Visualization of the samples for synthetic problems from ground truth. Top: Visualization of the samples generated with the proposed SBCG algorithm.}
\label{figure2}
\end{figure}

We next consider a few complicated two-dimensional  toy datasets, namely those from \cite{grathwohl2018scalable} with shapes of checkerboard, moons, pinwheel, and swissroll, respectively. For these distributions, we take the x-axis variable as $x$ and the y-axis variable as $z$. Same as what we did in previous example,  we first draw 50000 samples from target distributions for training the drift estimator. Then, for each given $z_i$ we generate an $\hat{x}_i$ from the SDE. The scatter plots of the generated 5000 pairs of $(x_i,z_i)$ are displayed at the top half of Figure \ref{figure2}. The scatter plots of 5000 samples drawn from the target distributions are displayed at the bottom half of Figure \ref{figure2}. Figure \ref{figure2} demonstrates that the empirical density of SBCG samples show a fairly good agreement with the ground truth density.

\subsubsection{Estimation of conditional mean and conditional standard deviation}\label{sec:mean}

In this section, we compare the performance of SBCG with several existing conditional density estimation methods in estimating the conditional mean and conditional standard deviation. The existing methods include the generative conditional distribution sampler (GCDS \cite{jiaocjasa}) and three popular conditional density estimation methods, i.e., the conditional kernel density estimation (CKDE \cite{fan2004}), nearest neighbor conditional density estimation (NNKCDE \cite{dalmasso2020conditional}), and flexible conditional density estimator (FlexCode \cite{izbicki2017converting}). Different from NNKCDE, CKDE and FlexCode, the SBCG and GCDS methods directly generate samples from the conditional density $\mu_{x^z}$ without estimating it. We consider the following three data generating models:
\begin{eqnarray}\nn
\text{Example 4}:&&    x = z_1^2 + \exp (z_2 + 0.25z_3) + \cos(z_4 + z_5) + \varepsilon,\\\nn
&& \varepsilon \sim N(0, 1),~\mathbf{z} =(z_1,\ldots,z_5)^{\rm T} \sim N(\mathbf{0}, \mathbf{I}_5),\\\nn   
\text{Example 5}:&& x = z_1^2 + \exp (z_2 + 0.25z_3) + z_4 - z_5 + (0.5 + 0.5z_2^2 + 0.5z_5^2)\varepsilon,\\\nn
&&\varepsilon \sim N(0, 1),~\mathbf{z} =(z_1,\ldots,z_5)^{\rm T} \sim N(\mathbf{0}, \mathbf{I}_5),\\\nn   
\text{Example 6}:&& x = \mathbb{I}_{\{U \leq 0.5\}} z_1 + \mathbb{I}_{\{U > 0.5\}} z_2,\\\nn
&&U \sim \mathcal{U}(0, 1)\, , z_1 \sim N(-Y, 0.25^2),~z_2 \sim N(Y, 0.25^2)\, , Y \sim N(0, 1).
\end{eqnarray}
Note that Example 4 represents a nonlinear model with additive error, Example 5 represents a model with an additive error whose variance depends on the predictors, and Example 6 represents a mixture of two normal distributions. For each model, we first generate 50000 joint samples of $(x,\mathbf{z})$ to train the neural network. Then we generate 2000 additional samples of $\mathbf{z}$ from the associated marginal distribution. For each $\bz$, we generate 200 conditional samples of $x$ using Algorithm \ref{alg2} based on the estimated drift term $\hat{\bu}^\star$ from the training step. The conditional mean and conditional standard derivation can be estimated using the 200 samples at given $\bz$. We implement SBCG using reference SDE (\ref{ref2}) with $\beta(t)=t$.  The neural networks used in the analysis is a fully connected network with 4 hidden layers with widths 32, 64, 64, 32. 

Denote ${\rm MSE}_1$ the mean squared difference between the estimated conditional mean and the true conditional mean at 2000 different $\bz$'s. Similarly, denote ${\rm MSE}_2$ the mean squared difference between the estimated conditional standard deviation and the true conditional standard deviation at 2000 different $\bz$'s. For each example, we repeat the process 100 times and report the summary of ${\rm MSE}_1$ and ${\rm MSE}_2$ over 100 replications for different methods in Table \ref{tab:simulation2}. Table \ref{tab:simulation2} shows that  (i) in terms of estimation error, the SBCG and GCDS methods outperform NNKCDE, CKDE and FlexCode in most cases, (ii) SBCG performs better than GCDS in most cases.

\begin{table}[H]
	\centering
	\begin{tabular}{ccccccc} \hline
& & SBCG& GCDS & NNKCDE & CKDE & FlexCode \\ \hline
		\multirow{2}{*}{E4}&${\rm MSE}_1$ &  \textbf{0.063}(0.029)&  0.259(0.015)  & 1.367(0.010) & 0.491(0.024) & 0.610(0.008) \\
		&${\rm MSE}_2$ &  \textbf{0.007}(0.001) &0.022(0.004) & 0.258(0.004) & 0.233(0.005) & 0.170(0.007) \\ \hline
&${\rm MSE}_1$ &  \textbf{0.295}(0.097)  &0.312(0.017)  & 4.668(0.046) & 1.707(0.060) & 2.408(0.063) \\
\multirow{-2}{*}{E5}&${\rm MSE}_2$ &  0.292(0.043) &\textbf{0.247}(0.012) & 0.793(0.008) & 0.857(0.017) & 2.384(0.602) \\ \hline
&${\rm MSE}_1$ &  0.012(0.004)&  0.016(0.003) & \textbf{0.004}(0.001) & 0.063(0.002) & 0.006(0.002) \\
\multirow{-2}{*}{E6}&${\rm MSE}_2$ & \textbf{0.006}(0.001)& 0.027(0.005) & 0.131(0.001) & 0.076(0.001) & 0.046(0.001) \\ \hline
	\end{tabular}
	\caption{Mean squared error of the estimated conditional mean (MSE$_1$), the estimated conditional standard deviation (MSE$_2$) and the corresponding simulation standard errors (in parentheses) summarized over 100 replications.
The smallest MSEs are in bold font.}
	\label{tab:simulation2}
\end{table}

\subsection{Wine quality data and abalone data} \label{subsec:wine}

In this section, we apply the proposed method to the wine quality dataset and abalone dataset from UCI machine learning repository \citep{ucidataset}. The wine quality dataset includes 6497 red and white vinho verde wine samples. The main purpose of this dataset is to rank the wine quality (discrete score between 0 and 10) using 11 chemical analysis measurements: fixed acidity, volatile acidity, citric acid, residual sugar, chlorides, free sulfur dioxide, total sulfur dioxide, density, pH, sulphates, and alcohol. All the measurements are continuous quantities. The abalone dataset contains 4177 rings of abalone and other physical measurements. The determination of the age of abalone is a time-consuming process which involves cutting the shell through the cone, staining it, and counting the number of rings through a microscope. On the other hand, the physical measurements \textit{sex, length, diameter, height, whole weight, shucked weight, viscera weight, shell weight} are easier to obtain. Therefore, the main purpose of abalone dataset is to use the physical measurements to predict the number of rings that determines the age. Except for the categorical variable \textit{sex}, all the other variables in this dataset are continuous. The variable \textit{sex} codes three groups: female, male and infant, since the gender of an infant abalone is not known. 

We take the score of wine quality in the wine quality data and the number of rings in the abalone data as the response denoted by $x$. Similarly, we take the 11 chemical analysis measurements in the wine quality data and the physical measurements in the abalone data as the covariate vector $\bz$. We randomly split the data into two parts, 90\% for training and the rest 10\% for testing. We estimate the conditional generator based on the training data and then evaluate it using the test data. Particularly, for each dataset, we first train the drift estimator on the training set, and then for each $\mathbf{z}_i$ from the test set, we generate $200$ $x$ samples denoted as $\{\hat{x}_i^{(j)}\}_{j=1}^{200}$ by Algorithm \ref{alg2}. We implement SBCG using reference SDE (\ref{ref2}) with $\beta(t)=\beta_{\text{min}}+(\beta_{\text{max}}-\beta_{\text{min}})t$ letting $\beta_{\text{min}}=1$ and $\beta_{\text{max}}=10$. The neural network is the same as the one used in the analysis in Section \ref{sec:visu}. Denote $\bar{\hat{x}}_i$ and $\hat{s}_i$ the sample mean and sample standard deviation of $\{\hat{x}_i^{(j)}\}_{j=1}^{200}$, respectively. Then the prediction intervals of response for given features $\mathbf{z}_i$ can be computed as 
\begin{align*}
\mathcal{C}_{i,1-\alpha}(\mathbf{z}_i)=\big\{x:\bar{\hat{x}}_i - z_{1-\alpha/2} \hat{s}_i \leq x \leq \bar{\hat{x}}_i + z_{1-\alpha/2} \hat{s}_i \big\}
\end{align*}
where $z_{1-\alpha/2}$ is the $(1-\alpha/2)$-quantile of standard normal distribution. Then we compute the coverage rate $\mathcal{C}_{i,1-\alpha}$ over the test data as
\[
{\rm CR}_{1-\alpha}=\frac{1}{N_{test}} \sum_{i=1}^{N_{test}} \mathbb{I}\{x_i\in\mathcal{C}_{i,1-\alpha}(\mathbf{z}_i)\}.
\]
With selecting $\alpha =$ 10\%, 5\% and 1\%,  Table \ref{tab:wine} reports the associated ${\rm CR}_{1-\alpha}$, which shows that ${\rm CR}_{1-\alpha}$ constructed from the SBCG based interval prediction methods are closer to the nominal level of $1-\alpha$ for both datasets.  

\begin{table}[htpb]
  \centering
   \caption{Associated ${\rm CR}_{1-\alpha}$ for two datasets with different selections of $\alpha$.}
   \bigskip
   \label{tab:wine}  
  \begin{tabular}{cccc}
    \toprule
    & $\alpha=0.1$ & $\alpha=0.05$ & $\alpha=0.01$ \\
    \midrule
    Abalone & 0.906 (0.030) & 0.943 (0.017) & 0.977 (0.013) \\
    Wine quality & 0.907 (0.010)& 0.947 (0.008) & 0.982 (0.005) \\
    \bottomrule
  \end{tabular}
\end{table}

\subsection{Image experiments}\label{subsec:mnist}

We now apply SBCG to the MNIST handwritten digits dataset \citep{lecun-mnisthandwrittendigit-2010} to illustrate that it can handle the models when either or both of $\bx$ and $\bz$ are high-dimensional. The dataset contains 60,000 images stored in $28\times 28$ matrices with gray color intensity from 0 to 1. Every image in the dataset belongs to one of the ten classes (digit 0 to 9). We use SBCG to generate images of specific categories by conditioning on class labels in section \ref{subsec:class_cond} and reconstruct the missing part of images in section \ref{subsec:impainting}.

\subsubsection{Generate images from labels}\label{subsec:class_cond}

\begin{figure}[ht!]
\centering
    \vspace{0cm}
		\begin{minipage}[t]{0.49\linewidth}
\includegraphics[width=1.02\textwidth]{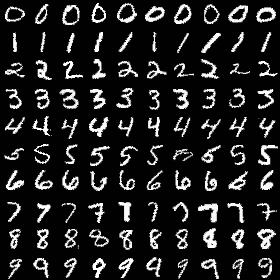}
		\end{minipage}
		\hspace{0.1cm}
		\begin{minipage}[t]{0.49\linewidth}
\includegraphics[width=1.02\textwidth]{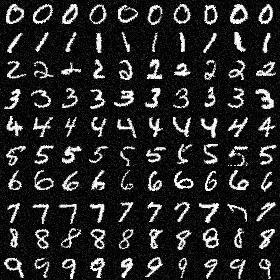}
		\end{minipage}
    \vspace{-0cm}
\caption{MNIST dataset: real images (left panel) and generated images given the labels (right panel).}
\label{figure3}
\end{figure}

In the problem of generating images of handwritten digits given labels from $\{0, 1, \ldots, 9\}$, the condition is a categorical variable representing one of the ten digits, and the response $\bx$ is a $28\times 28$ matrix representing the intensity values. For a given class label, we first use (\ref{drift0}) to compute the drift term $\bu^\star$ based on the subset of data from the MNIST dataset that have the corresponding class label. Then starting from a fix point, we employ the Euler-Maruyama discretization of the stochastic process (\ref{sde0}) to generate image samples.
Figure \ref{figure3} displays the real images randomly drawn from the training set (left panel) and synthetic images by our method (right panel).  Each row represents 10 images from the same label; and each column, from top to bottom, represents labels ranging from 0 to 9.  We see that the generated images are similar to the real images and  and it is hard to distinguish the generated ones from the real images. Also, there are some differences across each row, indicating the random variations in the generating process. Here, we implement SBCG using reference SDE (\ref{ref1}) with $\alpha(t)=1$.

\subsubsection{Image Inpainting}\label{subsec:impainting}

\begin{figure}[ht!]
\centering
    \vspace{0cm}
		\begin{minipage}[t]{0.31\linewidth}
\includegraphics[width=1.02\textwidth]{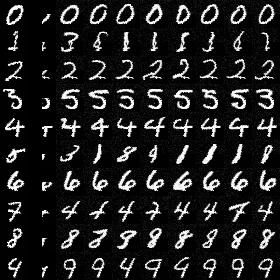}
		\end{minipage}
		\hspace{0.1cm}
		\begin{minipage}[t]{0.31\linewidth}
\includegraphics[width=1.02\textwidth]{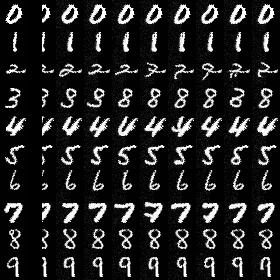}
		\end{minipage}
		\hspace{0.1cm}
		\begin{minipage}[t]{0.31\linewidth}
\includegraphics[width=1.02\textwidth]{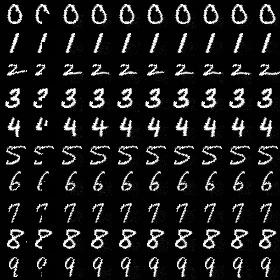}
		\end{minipage}
\caption{Reconstructed images given partial image in MNIST dataset. The first column in each panel consists of the true images, the second column in each panel consists of the associated conditions with part of image covered, the other columns give the constructed images. In the left panel, the right lower 1/4 of the image is given; in the middle panel, the right 1/2 of the image is given; in the right panel, 3/4 of the image is given}
\label{figure4}
    \vspace{-0.5cm}
\end{figure}

The goal of image inpainting is to reconstruct an image when part of the image is missing. It can be formulated as a conditional generative problem in which the observed part can be considered as the predictor $\bz$ and the missing part can be considered as the response $\bx$ which we need to sample. Suppose we want to reconstruct the image when just 1/4, 1/2, or 3/4 of its observation is given. For this problem, we first train the neural network to fit a MNIST training set with size 10000. We use networks of three layers of width 64 with SeLU activations. Then we prepare the associated condition $\mathbf{z}$ by using the same images as those from a testing set with part of the information covered. The first column of each panel in Figure \ref{figure4} displays the 10 images in the testing set with corresponding digit ranging from 0 to 9. The second column of each panel in Figure \ref{figure4} displays the associated conditions with part of image covered. The remaining 10 columns of each panel in Figure \ref{figure4} display the generated samples by Algorithm \ref{alg2} based on the given associated conditions in each row. 

In Figure \ref{figure4}, three plots from left to right corresponds to the situations when 1/4, 1/2, and 3/4 of an image are given. The reconstructed results show that the SBCG method are able to reconstruct all images correctly if 3/4 of the images are given. The digits “0”, “2”, "4", "6", "8", and “9” are easy to reconstruct. Even when only 1/4 of their images are given, SBCG can correctly reconstruct them. As the given area increases from 1/4 to 1/2 of the images, GCDS is able to reconstruct the images correctly for digits “1” and “7”. The digit "3" is more difficult and the reconstruction is only successful when 3/4 of the image is given. 

\section{Conclusion}\label{conclusion}

In this paper, we develop a SB-based generative method for sampling data from the conditional density. In contrast to the conditional score-based generative model (CSGM), our method is built based on the diffusion process that transform a fixed point to the target distribution in a finite time interval. Our method has the advantage over CDSB \citep{shi2022} in that it is a one-step simulation free algorithm which does not need to run iteration. The validity and accuracy of our method are accessed using numerical experiments on both synthetic and real datasets. Our numerical studies demonstrate that though our proposed method do not directly provide the conditional density estimation, the samples generated by this method can be effectively utilized to estimate the conditional density and  related statistical quantities, such as conditional mean and conditional standard deviation.

Certain generative models, such as GAN and flow-based models, are able to perform truncated or low temperature sampling to decrease the diversity of the samples while increasing the quality of each individual sample. In the CSGM framework, classifier-free guidance was developed in \cite{ho2022} to achieve the similar goal in attaining a trade-off between sample quality and diversity. The idea is to introduce a mask signal to randomly ignore the guidance and unify the learning of conditional and unconditional score networks. In the same spirit, our framework could be designed to first jointly train a conditional and unconditional diffusion model and then combine the resulting drift estimates to promote a balance between fidelity to the data and the entropy of the distribution, leading to more realistic and diverse samples. 

Conditional generative models represent a significant advancement in the field of generative modeling by providing a way to generate data that is not only realistic but also adheres to specific desired attributes. The ability to control the generative process makes them a versatile tool for a wide range of applications. For example, conditional generative learning offers an advantage over traditional statistical methods like regression by estimating the entire data distribution, not just the mean and standard deviation. Moreover, conditional generative learning provides an advantage in estimating multivariate density by capturing complex dependencies between variables, offering a more comprehensive representation of the data distribution compared to traditional methods. The integration of SB with conditional generative models provides a principled way to incorporate conditional information and can better handle noise and variability in the data, leading to more robust generation. Despite its promise, several challenges need to be addressed, such as the requirement for large amounts of labeled data, difficulties in stable training, and ensuring the diversity and quality of generated samples. Ongoing research aims to address these challenges through the development of advanced architectures, improved training methodologies, and exploration of more applications.

\appendix

\section{Proof of Proposition \ref{thm1}}\label{proof1}

\begin{proof}

The transition kernel for the stochastic process driven by SDE (\ref{ref1}) can be computed with equations (5.50) and (5.51) in \cite{Sarkka_Solin_2019} as
\begin{eqnarray}\nn
q(s,\bx_s,t,\bx_t)&=&N(\bx_t;\bx_s,[\alpha(t)-\alpha(s)]\bI_d).
\end{eqnarray}
Substitute this expression into (\ref{drift0}), we obtain
\begin{eqnarray}\label{ustar}
\bu^\star(\bx,t)&=&\frac{\alpha^\prime(t){\displaystyle\int}(\bx_1-\bx)f_t(\bx,\bx_1)\mu(d\bx_1)}{[\alpha(1)-\alpha(t)]{\displaystyle\int}f_t(\bx,\bx_1)\mu(d\bx_1)},
\end{eqnarray}
where 
\begin{eqnarray}\label{vef}
f_t(\bx,\bx_1)&=&\exp\left(\frac{\|\bx_1-\ba\|^2}{2[\alpha(1)-\alpha(0)]}-\frac{\|\bx_1-\bx\|^2}{2[\alpha(1)-\alpha(t)]}\right),
\end{eqnarray}
where $\|\cdot\|$ denotes the $L_2$-norm. Define conditional density of $\bx_t$ given $\bx_1$ as $\pi(\bx_t|\bx_1)\sim N(\bmu_{\bx_t|\bx_1},\sigma^2_{\bx_t|\bx_1})$, where
\begin{eqnarray}\nn
\bmu_{\bx_t|\bx_1}&=&\frac{\alpha(t)-\alpha(0)}{\alpha(1)-\alpha(0)}\bx_1+\frac{\alpha(1)-\alpha(t)}{\alpha(1)-\alpha(0)}\ba,\\\nn
\sigma^2_{\bx_t|\bx_1}&=&\frac{[\alpha(t)-\alpha(0)][\alpha(1)-\alpha(t)]}{\alpha(1)-\alpha(0)}.
\end{eqnarray}
Then (\ref{ustar}) can be rewritten as 
\begin{eqnarray}\nn
\bu^\star(\bx,t)&=&\frac{\alpha^\prime(t){\displaystyle\int}(\bx_1-\bx)\pi(\bx_t=\bx|\bx_1)\mu(d\bx_1)}{[\alpha(1)-\alpha(t)]{\displaystyle\int}\pi(\bx_t=\bx|\bx_1)\mu(d\bx_1)}\\\nn
&=&\frac{\alpha^\prime(t)}{\alpha(1)-\alpha(t)}{\displaystyle\int}(\bx_1-\bx)\frac{\pi(\bx_t=\bx,d\bx_1)}{\pi(\bx_t=\bx)}\\\nn
&=&\frac{\alpha^\prime(t)}{\alpha(1)-\alpha(t)}{\displaystyle\int}(\bx_1-\bx)\pi(d\bx_1|\bx_t=\bx)\\\label{ualt}
&=&\frac{\alpha^\prime(t)}{\alpha(1)-\alpha(t)}\mE_{\bQ_{\bx_1|\bx_t=\bx}}(\bx_1-\bx).
\end{eqnarray}
For any vector-valued function $\bu(\bx,t):\bR^{d}\otimes[0,1]\rightarrow\bR^{d}$, we have 
\begin{eqnarray}\nn
&&\mE_{\bQ_{\bx_1,\bx_t}}\left\|\frac{\alpha^\prime(t)}{\alpha(1)-\alpha(t)}(\bx_1-\bx_t)-\bu(\bx_t,t)\right\|^2\\\nn
&=&\mE_{\bQ_{\bx_1,\bx_t}}\left\|\frac{\alpha^\prime(t)}{\alpha(1)-\alpha(t)}(\bx_1-\bx_t)-\bu^\star(\bx_t,t)+\bu^\star(\bx_t,t)-\bu(\bx_t,t)\right\|^2\\\nn
&=&\mE_{\bQ_{\bx_1,\bx_t}}\left\|\frac{\alpha^\prime(t)}{\alpha(1)-\alpha(t)}(\bx_1-\bx_t)-\bu^\star(\bx_t,t)\right\|^2+\mE_{\bQ_{\bx_1,\bx_t}}\left\|\bu^\star(\bx_t,t)-\bu(\bx_t,t)\right\|^2\\\nn
&&+2\mE_{\bQ_{\bx_1,\bx_t}}\left\langle\frac{\alpha^\prime(t)}{\alpha(1)-\alpha(t)}(\bx_1-\bx_t)-\bu^\star(\bx_t,t),\bu^\star(\bx_t,t)-\bu(\bx_t,t)\right\rangle\\\label{ineq}
&\ge&\mE_{\bQ_{\bx_1,\bx_t}}\left\|\frac{\alpha^\prime(t)}{\alpha(1)-\alpha(t)}(\bx_1-\bx_t)-\bu^\star(\bx_t,t)\right\|^2,
\end{eqnarray}
for any $t\in[0,1]$. In deriving the last step, we have used the fact that 
\begin{eqnarray}\nn
&&\mE_{\bQ_{\bx_1,\bx_t}}\left\langle\frac{\alpha^\prime(t)}{\alpha(1)-\alpha(t)}(\bx_1-\bx_t)-\bu^\star(\bx_t,t),\bu^\star(\bx_t,t)-\bu(\bx_t,t)\right\rangle\\\nn
&=&\mE_{\bQ_{\bx_t}}\left\{\mE_{\bQ_{\bx_1|\bx_t}}\left\langle\frac{\alpha^\prime(t)}{\alpha(1)-\alpha(t)}(\bx_1-\bx_t)-\bu^\star(\bx_t,t),\bu^\star(\bx_t,t)-\bu(\bx_t,t)\right\rangle\right\}\\\nn
&=&\mE_{\bQ_{\bx_t}}\left\{\left\langle\frac{\alpha^\prime(t)}{\alpha(1)-\alpha(t)}\mE_{\bQ_{\bx_1|\bx_t}}(\bx_1-\bx_t)-\bu^\star(\bx_t,t),\bu^\star(\bx_t,t)-\bu(\bx_t,t)\right\rangle\right\}\\\nn
&=&0,
\end{eqnarray}
accoding to (\ref{ualt}). In (\ref{ineq}), the equality holds if and only if $\bu(\bx_t,t)=\bu^\star(\bx_t,t)$, therefore we prove that $\bu^\star(\bx_t,t)$ is the unique minimizer of (\ref{lse01}). 

Similarly, the transition kernel for the stochastic process driven by SDE (\ref{ref2}) can be computed as
\begin{eqnarray}\nn
q(s,\bx_s,t,\bx_t)&=&N(\bx_t;\bx_s e^{-\frac{1}{2}\int_s^t\beta(s^\prime)ds^\prime},[1-e^{-\int_s^t\beta(s^\prime)ds^\prime}]\bI_d).
\end{eqnarray}
Substitute this expression into (\ref{drift0}), we obtain
\begin{eqnarray}\nn
\bu^\star(\bx,t)&=&\frac{\beta(t)e^{-\frac{1}{2}\int_t^1\beta(s^\prime)ds^\prime}{\displaystyle\int}\left[\bx_1-\bx e^{-\frac{1}{2}\int_t^1\beta(s^\prime)ds^\prime}\right]f_t(\bx,\bx_1)\mu(d\bx_1)}{(1-e^{-\int_t^1\beta(s^\prime)ds^\prime}){\displaystyle\int}f_t(\bx,\bx_1)\mu(d\bx_1)},
\end{eqnarray}
where
\begin{eqnarray}\nn
f_t(\bx,\bx_1)&=&\exp\left(\frac{\|\bx_1-\ba e^{-\frac{1}{2}\int_0^1\beta(s^\prime)ds^\prime}\|^2}{2(1-e^{-\int_0^1\beta(s^\prime)ds^\prime})}-\frac{\|\bx_1-\bx e^{-\frac{1}{2}\int_t^1\beta(s^\prime)ds^\prime}\|^2}{2(1-e^{-\int_t^1\beta(s^\prime)ds^\prime})}\right).
\end{eqnarray}
Then following a similar procedure, we obtain that
\begin{eqnarray}\nn
\bu^\star(\bx,t)&=&\frac{\beta(t)e^{-\frac{1}{2}\int_t^1\beta(s^\prime)ds^\prime}}{1-e^{-\int_t^1\beta(s^\prime)ds^\prime}}\mE_{\mQ_{\bx_1|\bx_t=\bx}}\left[\bx_1-\bx e^{-\frac{1}{2}\int_t^1\beta(s^\prime)ds^\prime}\right],
\end{eqnarray}
which is the unique minimizer of (\ref{lse02}).
\end{proof}

\section{Proof of Proposition \ref{thm2}}\label{proof2}

\begin{proof}

The proof for the results of conditional case is a straightforward generalization of the proof for the unconditional case in Proposition \ref{thm1}. For (\ref{csde0}) with reference SDE (\ref{ref1}), first we can derive that 
\begin{eqnarray}\nn
\bu^\star(\bx,\bz,t)&=&\frac{\alpha^\prime(t)}{\alpha(1)-\alpha(t)}\mE_{\bQ^{\bz}_{\bx^\bz_1|\bx^\bz_t=\bx}}(\bx^\bz_1-\bx^\bz_t).
\end{eqnarray}
Then for any vector-valued function $\bu(\bx,\bz,t):\bR^{d_x}\otimes\bR^{d_z}\otimes[0,1]\rightarrow\bR^{d_x}$, we have 
\begin{eqnarray}\nn
&&\mE_{\tilde{\bQ}_t}\left\|\frac{\alpha^\prime(t)}{\alpha(1)-\alpha(t)}(\bx^\bz_1-\bx^\bz_t)-\bu(\bx^\bz_t,\bz,t)\right\|^2\\\nn
&=&\mE_{\mu_z}\mE_{\bQ^{\bz}_{\bx_1,\bx_t}}\left\|\frac{\alpha^\prime(t)}{\alpha(1)-\alpha(t)}(\bx^\bz_1-\bx^\bz_t)-\bu(\bx^\bz_t,\bz,t)\right\|^2\\\nn
&\ge&\mE_{\mu_z}\mE_{\bQ^{\bz}_{\bx^\bz_1,\bx^\bz_t}}\left\|\frac{\alpha^\prime(t)}{\alpha(1)-\alpha(t)}(\bx^\bz_1-\bx^\bz_t)-\bu^\star(\bx^\bz_t,\bz,t)\right\|^2,
\end{eqnarray}
for any $t\in[0,1]$. Therefore, $\bu^\star(\bx,\bz,t)$ is the unique minimizer of (\ref{ref1}) and thus we complete the proof of Proposition \ref{thm2}.
\end{proof}

\bibliographystyle{chicago} 
\bibliography{biblist}

\end{document}